# Memory Efficient Multi-Scale Line Detector Architecture for Retinal Blood Vessel Segmentation

Hamza Bendaoudi, Farida Cheriet and J. M. Pierre Langlois
Department of Computer & Software Engineering
Polytechnique Montréal
Montréal (Québec), Canada
{hamza.bendaoudi, farida.cheriet, pierre.langlois}@polymtl.ca

*Abstract*—This paper presents a memory efficient architecture that implements the Multi-Scale Line Detector (MSLD) algorithm for real-time retinal blood vessel detection in fundus images on a Zynq FPGA. This implementation benefits from the FPGA parallelism to drastically reduce the memory requirements of the MSLD from two images to a few values. The architecture is optimized in terms of resource utilization by reusing the computations and optimizing the bit-width. The throughput is increased by designing fully pipelined functional units. The architecture is capable of achieving a comparable accuracy to its software implementation but 70× faster for low resolution images. For high resolution images, it achieves an acceleration by a factor of 323×.

## I. INTRODUCTION

Estimates indicate that more than 439 million people around the world will be affected by diabetes by 2030 [1]. Eye diseases such as diabetic retinopathy (DR) and Glaucoma are a common complication of diabetes. To prevent the risk of developing such complications that can lead to blindness and visual loss, regular screening using retinal imaging is necessary. With the aim of making early detection of eye diseases using retinal images more accessible, automated detection is proposed as a solution with great potential to make regular examination available to underserved populations. To ensure large scale screening, telemedicine solutions have been proposed such as EyePACS [2]. This technology will allow patients with eye problems to be evaluated and monitored by a remotely located eye doctor and/or automatic screening systems. A telemedicine center will receive large numbers of images for patients from all around the world. Taking into consideration the number of images, and their high resolution, the processing time becomes an important factor. Retinal blood vessels detection is a key step in the process of eye disease detection. It allows to remove the vessels to analyse the lesions in CAD systems.

Recently, some hardware architectures have been proposed to accelerate the processing and to reduce the time required to detect retinal blood vessels. Hardware architectures implemented on FPGAs are proposed in [3] and [4]. Single Instruction Multiple Data (SIMD) architectures are also proposed in [5] and [6]. These implementations only consider images of low resolution. Graphical Processing Units (GPU) are also considered to accelerate the retinal image blood vessel segmentation in high resolution images [7] [8]. The previous cited implementations are able to achieve significant improvements over CPU implementations in terms of execution time.

Blood vessel segmentation performance is a crucial parameter. The implemented algorithms [3]-[8] do not consider the blood vessels at different scales which results in a miss of small vessels. In order to improve the small retinal blood vessel segmentation quality, several techniques and algorithms have been developed. Many authors have proposed to use supervised algorithms [9] [10] [11]. The major weakness of the supervised methods is that they are very dependent on the training datasets. These methods are not suitable for hardware implementation, because, once hard coded, any change or update of the classifier will require a new implementation and major modifications. Unsupervised methods have also been proposed. In this work we are interested especially on the line operator methods. Line operators, previously used in mammography, were modified and introduced first by Ricci et al. [11] and used later by several authors [12] [13]. Nguyen et al. [14] proposed a multi-scale line detector based algorithm for retinal blood vessel detection. The multi scale line detector is doing better than the basic line detector in [11]. In addition, the multi-scale is obtained by changing the length of a basic line detector, which is more suitable for hardware implementation than Gaussian pyramids in [13]. A complete review of the different methodologies for retinal blood vessel segmentation can be found in [15].

In this paper, we are targeting the multi-scale line detector (MSLD) for retinal image blood vessels detection. Since the MSLD works at multiple scales, the intermediate results need to be stored in memory, which increases the memory overhead. For this purpose, we propose a new architecture capable of solving the problem of memory requirements for the multi-scale line operator. Our solution is based on computation reuse and parallel implementation of the computations of each scale. The architecture is optimized in terms of resources utilization and throughput.

The remainder of this paper is organized as follows. Section II presents the multi-scale line detector for retinal image blood vessels detection. Section III gives a detailed description of the proposed architecture. Section IV discusses and evaluates the experimental results, and section V draws the conclusion.

## II. MSLD Algorithm Description

The MSLD algorithm was proposed by Nguyen el al. [14] as a generalized case of the line detector first proposed by Ricci et al [11]. The basic line detector is applied to the inverted green channel where the retinal blood vessels appear brighter than the background. For each pixel of the image, we consider a window of $W \times W$ pixels and the average gray level is computed as $I_{avg}^W$. We consider also twelve lines of length $W$ pixels oriented in 12 directions, centered on the pixel to process. The average gray level is computed along each line, and the maximum value is defined as $I_{max}^W$ and the line response at this pixel is then computed as:

$$R = I_{max}^W - I_{avg}^W \quad (1)$$

To improve the line detector for retinal blood vessel detection, Nguyen el al. [14] proposed the generalized line detector that works at multiple scales. The MSLD is based on varying the length of the aligned lines and (1) is redefined as:

$$R_W^L = I_{max}^L - I_{avg}^W \quad (2)$$

where $1 \leq L \leq W$, $I_{max}^L$ and $I_{avg}^W$ are defined as above. Line detectors at different scales are achieved by changing the value of $L$.

For each scale, we standardize the values of the raw response image to make them have zero mean and unit standard deviation distribution. The main purpose of the standardization is to achieve better contrast between the blood vessels and the retinal image background. The standardization is defined as:

$$R' = \frac{R - R_{mean}}{R_{Std}} \quad (3)$$

where $R'$ is the standardized response value, $R$ is the raw response value, $R_{mean}$ and $R_{Std}$ are the mean and standard deviation of the raw response values, respectively.

The response at each image pixel is the linear combination of the line responses of different scales, defined as:

$$R_{Combined} = \frac{1}{n_L + 1}\left(\sum_L R_W^L + I_{igc}\right) \quad (4)$$

where $n_L$ is the number of scales, $R_W^L$ is the response of the line detector at scale L and $I_{igc}$ is the value of the inverted green channel at the corresponding pixel.

## III. Proposed Memory-Efficient Architecture

This section presents the proposed architecture for the MSLD algorithm. It also describes the different steps of the system with more details, and it discusses the memory efficiency of the proposed architecture.

### A. System overview

The proposed system is implemented in a Zynq-7000 AP SoC that includes a dual-core ARM processing system (PS) with a 7-series Xilinx Programmable Logic fabric (PL) in a single device. The communication PS-PL is established using the Xillybus core [16]. Fig. 1 shows an overview of the proposed system for retinal blood vessel segmentation using MSLD.

### B. Image and mask loading

The first step to process the retinal image is the image and mask loading. The mask is a binary image that defines the ROI that corresponds to the retina. We use a binary image mask with the same size as the retinal image. Fig. 2 shows an example of an original image and its corresponding mask. Once the image and its mask are available to the Linux OS, a software program running on the ARM processor loads them and starts sending them to the PL at the same time. Only the green channel is processed.

### C. Raw response computation unit

The computation of the raw response for the different scales requires access to the image pixels corresponding to the window centered on the pixel to be processed. For this purpose, a shift register of length $(W - 1) \times Ncols + W$ is used, where $Ncols$ is the number of columns of the retinal image. Since the MSLD works on the inverted green channel, the pixels of the image are subtracted from 255 before being pushed to the line buffer. All the pixels inside the window are needed to compute the mean value of the window. Only the pixels that correspond to the line detector for the twelve different orientations are fed to the modules that compute the raw response.

For each scale, we compute the mean value of the intensity level of the pixels along each line for the 12 orientations. The maximum mean value among the different lines is sorted and the raw response of the scale is the subtraction of the maximum value from the mean value of the window.

Fig. 3 shows the architecture of the Line Response Computation Module (LRCM). This figure shows how the pixels of one line (from a window of $11 \times 11$ pixels) are arranged to realize a fully pipelined adder tree. The output of the scale 1 is used to compute the result for scale 2 and so on. In this way, the computations are reused to reduce the number of functional units. At each stage of the pipeline, the adders are tailored for the exact word-length of the outputs. The output of each scale is multiplied by the corresponding reciprocal coefficient to realize the division by the number of pixels and then compute the mean value. By reusing the computations and word-length optimization, resource consumption is reduced. Pipelining reduces the critical path and increases throughput.

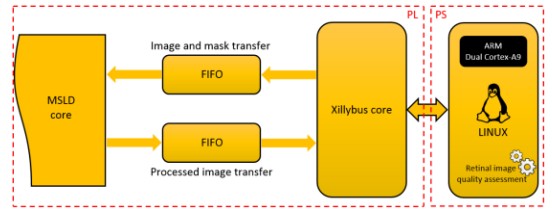

Fig. 1. Overview of the Zynq-based system for retinal blood vessel detection.

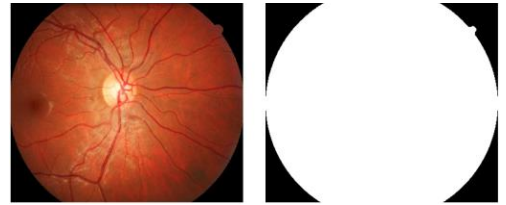

Fig. 2. Original image and the corresponding mask.

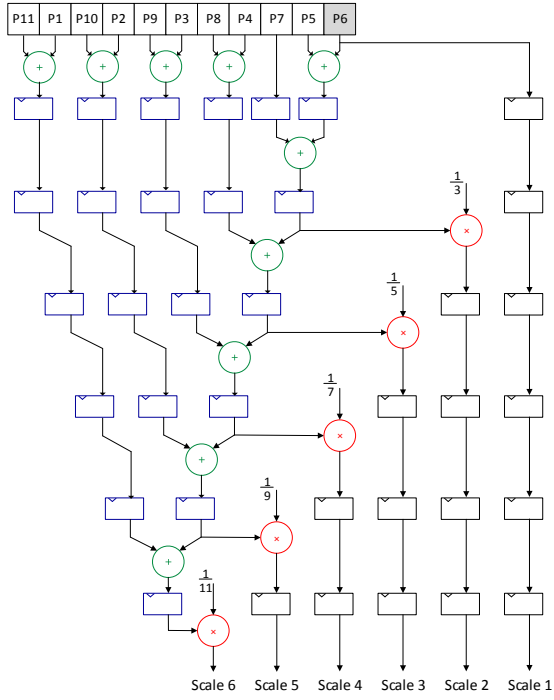

Fig. 3. Line response computing module

For each pixel of the window, the line response is computed for the twelve different orientations and at different scales. The goal behind this operation is to compute one raw response at each scale. For this purpose, 12 LRCM blocks are used as shown in Fig. 5. The outputs of these blocks are routed to the six Raw Response Computation Module RRCM (the number of scales equals to $(W + 1)/2$, so for an $11 \times 11$ window there are have six scales). The RRCM blocks are responsible for computing the raw response for one scale. A RRCM block has 12 inputs that correspond to one scale but from different lines and one input for the mean value inside the window. As can be seen in Fig. 5, the RRCM for the scale 1 has as input the $S_1$ outputs of the twelve LRCM blocks and the mean value of the window (in red color).

Fig. 4 shows the architecture of the RRCM block. The response of each scale is the difference between the mean value of the pixels inside the window and the maximum response of the 12 lines. Sorting the maximum response of the 12 lines is realised by comparing the inputs two by two in parallel and in a pipelined way. The mean value computation module is designed as a tree to be fully pipelined with its word-length optimized to increase the frequency and reduce the computation resources, respectively.

*D. Mean and standard deviation values computation*

In parallel with the raw response computation, the mean and standard deviation values are also computed in a streaming way. To compute the mean value, the intensity values of the raw responses that correspond to the ROI are accumulated. At the same time, the number of accumulated values is counted since the ROI is circular and the number of pixels inside it is unknown. The ROI is defined by the mask by a one bit signal. The mean value is computed according to (5).

$$m = \frac{1}{N}\sum x_i \quad (5)$$

At the end of the image, the sum of the pixel values is divided by the number of pixels to get the final mean value. The standard deviation value is computed according to (6).

$$\sigma^2 = \frac{1}{N}\sum x_i^2 - m^2 \quad (6)$$

For this purpose, the square value of the intensity value of the raw responses is computed and then accumulated when the pixel is inside the ROI. At the end of the image, the accumulated value is divided by the number of pixels, then, the variance value is computed. To compute the standard deviation value the square root of the variance is computed as shown in Fig. 6. The red part shows the computation of the number of pixels inside the ROI. The blue part shows the mean value computation, and the green part shows the standard deviation value computation. The mean and standard deviation values are computed for each scale. Until this step, all the computations are done in a streaming manner and in parallel for all the scales without saving the raw responses. Instead we save the mean and standard deviation values for all the scales.

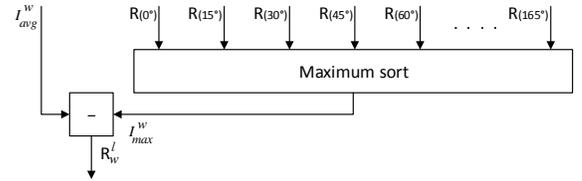

Fig. 4. Raw response computation module architecture.

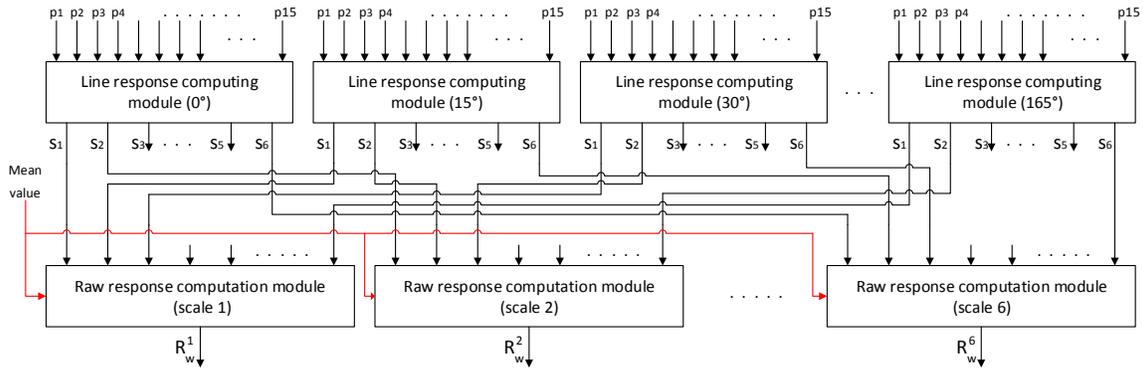

Fig. 5. Raw response computation at the different scales.

## E. Standardized and combined response computation

The mean and standard deviation are computed to standardize the raw responses. Since the standardization step necessitates the raw response and the mean and standard deviation values at the same time, this a challenging problem since the FPGA does not have sufficient on-chip memory to store the raw responses while computing the mean and standard deviation values. To overcome this problem, we propose to save the mean and standard deviation values for the different scales, which results in very low memory requirement in comparison with the raw responses saving requirements. Once the mean and standard deviation values are computed and stored, the computation of the raw responses is restarted for a second time without re-computing the mean and standard deviation values since they were already computed and stored.

For this second pass, each processed pixel for all the scales will be standardized and combined on the fly according to (3) and (4). Fig. 7 shows the principle of re-computing the raw responses to avoid saving them and to reduce the memory requirements. As can be seen in Fig. 7, in the first pass the raw responses, the mean and standard deviation values are computed. The objective is to avoid saving the raw responses since the on-chip memory is not sufficient to save such a large amount of data (for a 11×11 pixels window, 6 scales are handled in parallel, and the memory requirements are equal to 6 times the original image if stored).

Fig. 8 shows the architecture of the standardized and combined response computation module. For this purpose, we use 6 standardization modules, one for each scale. The inputs of the standardization modules are the newly computed raw response of the corresponding scale, the mean and standard deviation values.

## F. Operations scheduling

Operations scheduling is a vital task for the functioning of the entire system. Since the number of inputs of each scale is variable, the number of pipeline stages is also variable. For scale 1, only the center pixel of the window is needed, while scale $n$ needs $W$ bits to compute the line response. As shown in Fig. 3, the pixel values are added two by two with full pipeline. The scale 1 response is computed in the first stage of the pipeline, the scale 2 response is computed after two pipeline stages and so on for the other scales responses.

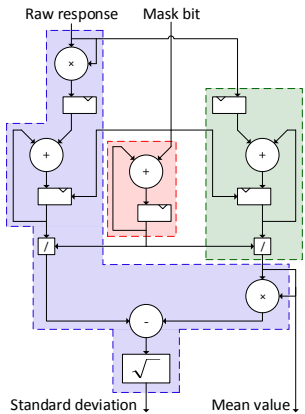

Fig. 6. Mean and standard deviation values computation module.

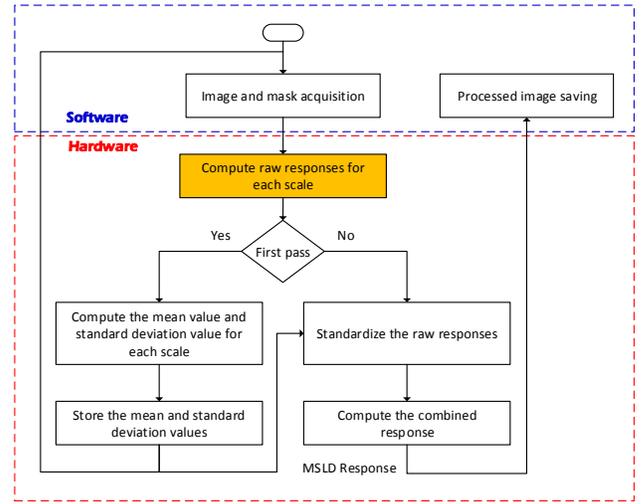

Fig. 7. Principle of re-computing raw responses to avoid saving them.

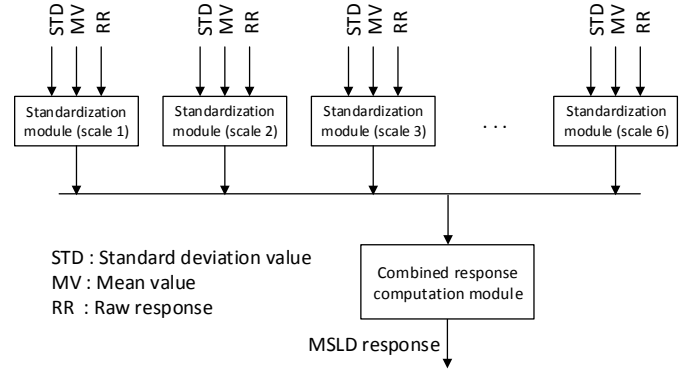

Fig. 8. Combined response computation architecture.

To properly compute the raw responses at each scale, the outputs of the LRCM module must be synchronized and sent at the same time to the RRCM modules. For this purpose, registers are added to the outputs of the scales to balance the pipeline and to delay the first computed responses to get them out at the same time with result of the scale n (largest scale). As can be seen in Fig. 3, the added registers (in black), for example, the scale 1 response is delayed by six clock cycles using six registers. For scale 2, four registers are added, and thus, the outputs are fully synchronized.

This is not the only place where registers are added to synchronize the outputs. The RRCM module needs the mean value of the window, where the window contains a large number of pixels ($W^2$ pixels). The number of stages of the RRCM module and the LRCM module is different, and then the first computed outputs should be delayed to meet the outputs of the second module. This is achieved by adding the necessary number of registers to balance the two pipelines.

## G. Memory efficient architecture

Fig. 9 shows the CPU and the custom parallel implementations details. For the CPU implementation, once the image and mask are acquired, the raw response computation can be started. For each value (pixel) of the raw response, the computation of the mean value and of the standard deviation value is started according to (5) and (6). The raw response for scale $i$ must be stored in the internal memory before the standardization step. The standardization requires the value of the raw response at each pixel, the mean value and the standard deviation for the scale $i$. The standard response is then computed according to (3). At the same time, the combined response is computed by accumulating the standardized responses. Consequently we store the new combined response once a new scale is processed. The final combined response is obtained once the last scale is processed.

The major weakness of the software implementation of the MSLD algorithm is the memory intensive nature of the algorithm. The MSLD algorithm works at multiple scales. The CPU does not have the possibility to handle a large parallelism to process all the scales in parallel, and, thus, the processing is done in a sequential manner. For each scale, the intermediate results (the raw response and the combined response) must be stored in memory. The amount of memory necessary is thus equal to 2× size of the original image as shown in Fig. 9.

The custom parallel implementation does not require to store the intermediate responses of the different scales. Instead, it requires to compute the raw responses twice. For the first run, the mean and standard deviation values are computed and stored. In the second run, the raw responses are computed again which allows the standardization of the raw responses on the fly using the stored mean and standard deviation values without the need to re-compute them. The custom parallel implementation takes advantage of the parallelism allowed. A fully parallel data path is a suitable solution to handle the different scales and to re-compute the raw responses twice very swiftly. The custom parallel implementation shrinks the memory requirements from 2 *images* to 2 × number of scales *values*.

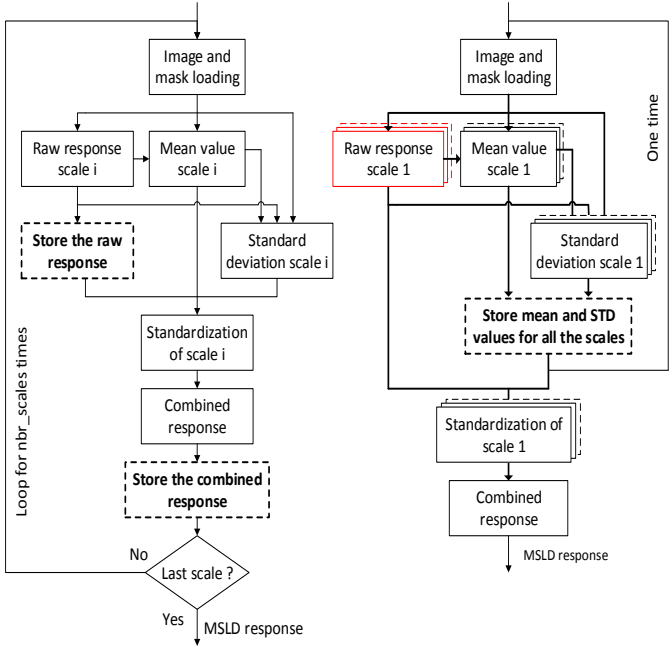

Fig. 9. CPU and custom parallel architecture implementation details.

## IV. RESULTS AND DISCUSSION

This section presents the implementation results of the MSLD algorithm for retinal blood vessel detection. The MSLD algorithm was implemented on CPU and FPGA. The CPU implementation was realized in C++ and implemented on an Intel i7-2600 CPU @ 3.4GHz, 16 GB of RAM. The FPGA implementation was coded in VHDL and implemented on Zynq®-7000 SoC XC7Z020-CLG484-1.

Fig. 10 shows a sample of an original retinal image with its binarized MSLD response. To evaluate the blood vessel detection performance, we use the publically available DRIVE databases collected by Staal et al. [9]. To quantify the blood vessel segmentation performance, Area Under the Curve (AUC), Sensitivity (SE), Specificity (SP) and Accuracy (ACC) metrics are computed.

Table I shows the quality measures of blood vessel detection for DRIVE database for the CPU and FPGA implementations. As can be seen in Table I, the quality measures of the CPU implementation are comparable. The difference in results is due to the fixed point computations precision of the FPGA implementation compared to the floating point full precision of the CPU. The FPGA implementation functional units are optimized in terms of word-length. The integer part of the signals are all set to the right value. The fractional part is set to 18 bits for DRIVE database images. The number of bits of the fractional part is set with respect to certain precision of computation.

For low resolution images such as the images of DRIVE database, the optimal parameter $W$ of the MSLD algorithm is $W = 15$. Eight scales are considered (for $L$ from 1 to 15 with a step of 2). For the CPU implementation, the loop is executed 8 times and two images of $565 \times 584$ pixels are stored in memory, while the FPGA implementation need to store $2 \times 8$ values. For high resolution images of $3504 \times 2336$ pixels images, the loop is executed 21 times, and 2 images of $3504 \times 2336$ pixels are stored in memory. For the FPGA implementation, only $2 \times 21$ values are stored.

Table II. shows the FPGA and CPU implementations performance measures. For the first test case ($565 \times 584$ pixels images) the MSLD parameter $W = 15$. For the second test case ($3504 \times 2336$ pixels images), the parameter $W = 41$. As we can see in the table, for low resolution images, the execution time for FPGA is 0.014 s with a throughput of 71.428 frames/s (f/s). The execution time for CPU is 0.988 s with a throughput of 1.012 f/s. The FPGA implementation is 70.57 times faster than the CPU implementation for low resolution images.

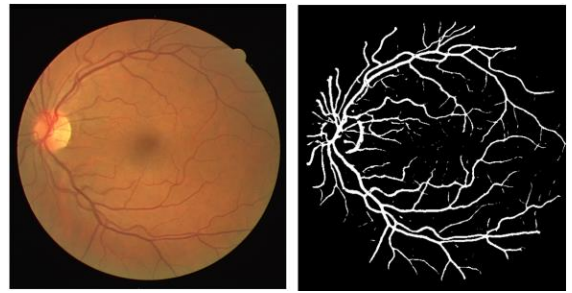

Fig. 10. Sample images. Original image and binarized MSLD.

Table I. Performance measures of blood vessel segmentation for DRIVE database images – CPU and FPGA implementations.

|      | AUC    | SE     | SP     | ACC    |
|------|--------|--------|--------|--------|
| CPU  | 0.9358 | 0.7425 | 0.9701 | 0.9407 |
| FPGA | 0.9363 | 0.7527 | 0.9643 | 0.9370 |

Table II. FPGA versus CPU implementation performances

| Image size | Platform | Time (s) | Throughput (f/s) | Speed Up |
|------------|----------|----------|------------------|----------|
| 565×584    | CPU      | 0.988    | 1.012            | 1        |
|            | FPGA     | 0.014    | 71.428           | 70.5 ×   |
| 3504×2336  | CPU      | 144.611  | 0.006            | 1        |
|            | FPGA     | 0.447    | 2.232            | 323.5 ×  |

Table III. Logic utilisation for the MSLD algorithm for DRIVE database images.

| FPGA Resources    | Used  | Available | Utilisation |
|-------------------|-------|-----------|-------------|
| LUTs              | 10427 | 53200     | 20%         |
| Flip-Flops        | 7498  | 106400    | 7%          |
| DSP Blocs         | 110   | 220       | 50%         |
| Maximum Frequency | 60.443 MHz |      |             |

For higher resolution images, the generated circuit consumes more resources than the available in the Zynq XC7Z020-CLG484-1. For this reason, our circuit is regenerated for the Zynq 7Z045-FFG900-2 FPGA. The execution time of the FPGA is 0.447 s with a throughput of 2.232 f/s. The execution time for the CPU is 144.611 s with a throughput of 0.006 f/s. The FPGA implementation is 323.5 times faster than the CPU implementation for high resolution images.

## V. CONCLUSION

In this paper, we proposed a hardware architecture for the MSLD algorithm. The proposed architecture is optimized to minimize the FPGA resources consumption and to increase the throughput. The architecture benefits from the massive parallelism offered by the FPGAs to reduce the memory requirements of the MSLD algorithm. Our FPGA implementation is drastically effective in terms of memory. Instead of saving several images for the software implementation, our FPGA implementation requires to save few values to give comparable performances in terms of blood vessels detection accuracy with higher throughput. The FPGA implementation is 70× and 323× faster than the software implementation for low and high resolution images, respectively.